\documentclass{article}
\usepackage{spconf,amsmath,graphicx,hyperref}
\usepackage[T1]{fontenc}
\usepackage{cite}
\usepackage{wrapfig}
\usepackage{amsfonts}
\usepackage{algorithmic}
\usepackage{graphicx}
\usepackage{multirow}
\usepackage{nicematrix}
\usepackage{booktabs} 
\usepackage[normalem]{ulem} 
\usepackage{lipsum} 
\usepackage{subcaption}

\usepackage[utf8]{inputenc}

\title{Variance \& GREEDINESS: A COMPARATIVE STUDY OF METRIC-LEARNING LOSSES}

\name{Donghuo Zeng \qquad Hao Niu \qquad Zhi Li \qquad Masato Taya}
\address{KDDI Research, Inc., Japan}

\begin{document}
%
\maketitle
\begin{abstract}
Metric learning is central to retrieval, yet its effects on embedding geometry and optimization dynamics are not well understood. We introduce a diagnostic framework, VARIANCE (intra-/inter-class variance) and GREEDINESS (active ratio and gradient norms), to compare seven representative losses, i.e., Contrastive, Triplet, N-pair, InfoNCE, ArcFace, SCL, and CCL, across five image-retrieval datasets. Our analysis reveals that Triplet and SCL preserve higher within-class variance and clearer inter-class margins, leading to stronger top-1 retrieval in fine-grained settings. In contrast, Contrastive and InfoNCE compact embeddings are achieved quickly through many small updates, accelerating convergence but potentially oversimplifying class structures. N-pair achieves a large mean separation but with uneven spacing. These insights reveal a form of efficiency-granularity trade-off and provide practical guidance: prefer Triplet/SCL when diversity preservation and hard-sample discrimination are critical, and Contrastive/InfoNCE when faster embedding compaction is desired.
\end{abstract}

\section{Introduction}
Metric learning embeds inputs so that geometric proximity reflects semantic similarity, enabling nearest-neighbor retrieval~\cite{sohn2016improved, zeng2020deep, wang2019multi, zeng2024anchor, zeng2023two, zeng2026learning, zeng2022complete, zeng2025metric} and downstream~\cite{Manmatha2017SamplingMI, schroff2015facenet, li2015feature}. Popular supervised objectives — contrastive~\cite{hadsell2006dimensionality}/InfoNCE-style softmax losses and margin-based triplet formulations — both push class separation but differ in how they allocate gradient effort across samples. These differences in update patterns produce distinct embedding geometries (how compact or dispersed classes become) and different optimization dynamics (how rapidly training plateaus on easy samples)~\cite{ghojogh2020fisher, musgrave2020metric, hermans2017in, wang2019multi}. Understanding these effects is crucial for retrieval tasks where preserving subtle intra-class structure matters (fine-grained retrieval) or where rapid, stable compaction is preferred (coarse retrieval).

This paper studies two complementary axes of analysis. First, \textbf{VARIANCE} diagnostics quantify the geometry of final embeddings via intra-class and inter-class means/variances (Table~\ref{tab:mean_vars}). Second, \textbf{GREEDINESS} diagnostics characterize optimization behavior using the active-sample ratio (fraction of nonzero losses per batch), and gradient-norm statistics (Fig.~\ref{fig:greediness_metrics}). We apply this suite to seven supervised losses, i.e., Contrastive~\cite{hadsell2006dimensionality}, Triplet~\cite{schroff2015facenet, musgrave2020metric}, N-pair~\cite{sohn2016improved}, InfoNCE~\cite{oord2018representation}, ArcFace~\cite{Jiankang2018} Supervised contrastive loss (SCL~\cite{khosla2020supervised}), and Center contrastive loss (CCL~\cite{cai2023center})), across five datasets chosen for diversity in scale and semantic granularity.

Our empirical results reveal a consistent pattern: triplet loss and SCL maintain greater within-class variance while enforcing clear inter-class margins, which benefits top-1 retrieval on fine-grained datasets. In contrast, contrastive/ InfoNCE exhibits rapid, many-small-update dynamics, leading to quick intra-class compaction that favors coarse-grained retrieval but may obscure fine distinctions. N-pair loss often achieves strong centroid separation yet shows high inter-class variance, reflecting uneven class spacing and potential nearest-neighbor retrieval failures. These relationships between embedding geometry, training dynamics, and retrieval performance are summarized in Table~\ref{tab:retrieval_result} and discussed throughout the paper. 

Our main contributions are threefold: (1) \textbf{Methodological Framework}: We introduce the VARIANCE–GREEDINESS diagnostic suite for analyzing metric-learning objectives. (2) \textbf{Conceptual Discovery}: We conduct a systematic empirical comparison of seven losses on five image retrieval datasets, we identify a fundamental efficiency-granularity trade-off. We demonstrate that "greedy" optimization (e.g., in InfoNCE) facilitates rapid cluster compaction but can compromise the intra-class diversity preserved by margin-based losses (e.g., Triplet). (3) \textbf{Practical Guidelines}: We provide actionable selection criteria to match loss functions with task demands: recommending Triplet/SCL for fine-grained retrieval where diversity preservation is critical, and Contrastive/InfoNCE for scenarios prioritizing rapid convergence and high embedding density.
\vspace{-26pt}
\section{Methods}
\subsection{Variance diagnostics}
To quantify embedding geometry, we measure intra- (class dispersion) and inter-class variances (centroid separation):

\begin{equation}
 \begin{aligned}
\sigma_{\text{intra}}^2 = \frac{1}{C}\sum_{c=1}^C \frac{1}{N_c}\sum_{i \in I_c}\|z_i - \mu_c\|^2, 
\quad  \\
\sigma_{\text{inter}}^2 = \frac{1}{C(C-1)} \sum_{c \neq c'} \|\mu_c - \mu_{c'}\|^2,
\end{aligned}
\end{equation}

where $C$ is the number of classes, $N_c$ is the number of samples in class $c$, $\mu_c$ is its centroid, and $z_i = f(x_i)$, the output of the neural networks. Intra-class variance reflects cluster spread, while inter-class variance reflects class separation.

\subsection{Optimization greediness}
We define \emph{greediness} as the tendency of a loss to keep optimizing constraints that are already satisfied, leading to excessive compaction or even dimensional collapse~\cite{Jing2021UnderstandingDC}. We quantify greediness with two diagnostics:

\subsubsection{Active Ratio}
The \textit{active ratio} measures the fraction of samples in a batch that contribute to a non-zero loss, i.e., those for which the margin or classification constraint is not satisfied~\cite{Jing2021UnderstandingDC}.

\textit{Pair/triplet-based losses} (Contrastive, Triplet, N\-Pair, InfoNCE, SCL): A sample or triplet is active when its distance-based inequality is violated. For example, in triplet loss, $(a,p,n)$ is active if $d(a,p) + m > d(a,n)$. The active ratio is the number of such pairs/triplets divided by the batch size.
\textit{ArcFace:} A sample is active if its margin-augmented logit for the correct class is not the highest among all classes.  
\textit{CCL:} A sample is active if it is closer to the center of another class than its own, that is, $d(z_i, \mu_{y_i}) > \min_{c \neq y_i} d(z_i, \mu_c)$.
The active ratio is the fraction of samples with nonzero loss per batch~\cite{Manmatha2017SamplingMI}, which is measured by
\( = \frac{|\{(x,y) \in P \cup N : \mathcal{L}(x,y) > 0\}|}{|\text{Batch}|}\). This quantifies how many samples continue to drive learning.  

\subsubsection{Gradient norm}
The overall $\ell_2$ norm of parameter gradients after backpropagation, $\|\nabla \mathcal{L}\|_2 = \sqrt{\sum_j \|\nabla_{\theta_j} \mathcal{L}\|_2^2}$. Contrastive/InfoNCE typically yield high active ratios but low gradient norms, corresponding to many small updates across the batch~\cite{musgrave2020metric, Manmatha2017SamplingMI}. Triplet and SCL show the opposite pattern: lower active ratios but larger gradient norms, concentrating learning on harder examples~\cite{hermans2017in}. These contrasting behaviors underpin the convergence speed versus variance-preservation trade-offs analyzed in Section~\ref{sec:greediness_in_optimization}.

\section{Experimental Framework}
\begin{table}[h]
\centering
\small
\caption{Statistics of the benchmark datasets}
\begin{tabular}{lcc}
\toprule
\textbf{Dataset} & \textbf{Images (Classes)} & \textbf{Train/Test} \\
\midrule
CIFAR-10 & 60,000 (10) & 50,000/10,000 \\
Car196 & 16,185 (196) & 8,144/8,041 \\
CUB-200 & 11,788 (200) & 5,994/5,794 \\
Tiny-ImageNet & 120,000 (200) & 100,000/10,000 \\
FashionMNIST & 70,000 (10) & 60,000/10,000 \\
\bottomrule
\end{tabular}
\vspace{-15pt}
\label{tab:dataset_stats}
\end{table}

\subsection{Datasets}
We evaluate metric learning methods on five diverse image retrieval datasets, Table~\ref{tab:dataset_stats} shows each varying in size, number of classes, and train/test splits. 
\textit{CIFAR-10} is standard benchmarks for object recognition with low-resolution color images, differing in the number of classes. \textit{Car196} focuses on fine-grained car model classification, while \textit{CUB-200} targets bird species identification, both with variable-sized images.
\textit{Tiny-ImageNet} is a subset of ImageNet with smaller images across a wide range of object categories. \textit{FashionMNIST} contains grayscale clothing images for fashion item classification.
\vspace{-27pt}
\subsection{Settings}
\textit{Model architecture.} We use a frozen Vision Transformer backbone (ViT-B/32)~\cite{clipvit} as the feature extractor. The backbone output (768-D) is passed through a two-layer projection head with Tanh activations: $768 \!\rightarrow\! 512 \!\rightarrow\! 128$. Dropout (rate = 0.15) is applied between the two fully-connected layers. The final 128-D embedding is L2-normalized and used for all retrieval experiments. \textit{Optimization.} All models are trained with the Adam optimizer (initial learning rate $=1\times10^{-4}$, weight decay $=1\times10^{-5}$). We train for 100 epochs with a batch size of 512, and the fixed learning-rate schedule is used. \textit{Distance metric and losses.} Because the final embeddings are L2-normalized, we adopt cosine similarity (and report cosine distance $d_{\cos}=1-\langle z_i, z_j\rangle$) as the canonical pairwise measure for retrieval and for all VARIANCE diagnostics (intra-/inter-class means/variances). For training: (1) InfoNCE and Supervised Contrastive (SCL) are implemented using the dot-product (equivalent to cosine similarity on normalized embeddings) and temperature $\tau=0.07$. (2) Contrastive and Triplet objectives operate on euclidean distance with margin $m=1.0$; we found $m=2.0$ to be a stable default for our setup and tuned on validation splits. (3) N-pair and CCL use their standard formulations but rely on cosine similarity for pairwise terms; for CCL the center-regularizer weight is set to $\lambda_c = 10$. (4) All embeddings are L2-normalized prior to computing loss terms and nearest-neighbor retrieval. \textit{Diagnostics and visualization.} During training we log loss curves, active ratio (fraction of non-zero losses per batch), and gradient-norm statistics to analyze optimization dynamics. Final embeddings are visualized with t-SNE for qualitative inspection, and we report VARIANCE diagnostics (intra-/inter-class means and variances) in Table~\ref{tab:mean_vars} computed with cosine distance.

\begin{table*}[t]
\small
\centering
\setlength{\tabcolsep}{3pt} 
\caption{Statistics of intra- and inter-class variances ($\mu$ and $\sigma^2$) across five benchmark datasets}
\resizebox{\textwidth}{!}{
\begin{tabular}{l|cccc|cccc|cccc|cccc|cccc}
\toprule
\multirow{2}{*}{Loss}  & \multicolumn{4}{c}{CIFAR-10} & \multicolumn{4}{c}{CARS196} & \multicolumn{4}{c}{CUB-200} & \multicolumn{4}{c}{Tiny-ImageNet} & \multicolumn{4}{c}{FashionMNIST} \\ 
\cmidrule(lr){2-5} \cmidrule(lr){6-9} \cmidrule(lr){10-13} \cmidrule(lr){14-17} \cmidrule(lr){18-21} 
& \multicolumn{2}{c}{Intra-} & \multicolumn{2}{c}{Inter-} & \multicolumn{2}{c}{Intra-} & \multicolumn{2}{c}{Inter-} 
& \multicolumn{2}{c}{Intra-} & \multicolumn{2}{c}{Inter-} & \multicolumn{2}{c}{Intra-} & \multicolumn{2}{c}{Inter-} 
& \multicolumn{2}{c}{Intra-} & \multicolumn{2}{c}{Inter-}  \\ 
\cmidrule(lr){2-3} \cmidrule(lr){4-5} \cmidrule(lr){6-7} \cmidrule(lr){8-9} \cmidrule(lr){10-11} \cmidrule(lr){12-13} \cmidrule(lr){14-15} \cmidrule(lr){16-17}\cmidrule(lr){18-19} \cmidrule(lr){20-21} 
& $\mu$ & $\sigma^2$ & $\mu$ & $\sigma^2$ & $\mu$ & $\sigma^2$ & $\mu$ & $\sigma^2$ 
& $\mu$ & $\sigma^2$ & $\mu$ & $\sigma^2$ & $\mu$ & $\sigma^2$ & $\mu$ & $\sigma^2$
& $\mu$ & $\sigma^2$ & $\mu$ & $\sigma^2$  \\
\midrule
Contrastive   
    & 0.0656 &0.0713 &0.4790 &0.3402
    & 0.4008 & 0.0672  & 0.7221 & 0.2030
    & 0.3871 &0.0843 &0.8022 &0.1590
    & 0.3481 &0.0794 &0.6882 &0.1156
    & 0.2788 & 0.1629 & 1.2110 & 0.2208  \\
Triplet       
    &0.1435 &0.0601 &1.3653 &0.1327
    & 0.3861 & 0.1115 & 0.9580 &0.3297
    & 0.3569 &0.0978 &1.0590 &0.2460
    & 0.4432 & 0.0997 & 1.0397 & 0.1238
    & 0.2941 & 0.1675 & 1.2081 & 0.2025  \\
N-pair       
    & 0.0931 &0.0302 &1.4095 &0.1623
    & 0.3049 & 0.1604 & 1.0979 &0.4255 
    &0.2435 &0.1330 &1.1749 &0.3522
    &0.1815 &0.0997 &1.2350 &0.3367 
    & 0.0937 &0.0577 &1.3789 &0.2962  \\
InfoNCE        
    & 0.1091 &0.0526 &1.4055 &0.0330
    & 0.5845 & 0.1338 &0.8673 &0.1835
    & 0.5238 &0.1516 &0.9477 &0.1637
    &0.4134 &0.1300 &1.0796 &0.1014
    & 0.2661 &0.1301 &1.2536 &0.1379 \\
ArcFace         
    & 0.0000 &0.0000 &0.0079 &0.0030
    & 0.0078 &0.0145 & 0.1806 & 0.1400 
    & 0.0656 &0.0713 &0.4790 &0.3402
    & 0.0000 & 0.0000 & 0.0044 & 0.0016 
    & 0.0000 &0.0000 &0.0048 &0.0023 \\
SCL
    & 0.1234 &0.0327 &1.3573 &0.1265
    & 0.3634 &0.0910 &0.9088 &0.2719
    & 0.3937 &0.0948 &0.9541 &0.1960
    & 0.0485 &0.0696 &0.3564 &0.1935
    & 0.1085 &0.0458 &1.3414 &0.2655  \\
CCL
    & 0.0346 &0.0140 &0.6048 &0.3057
    & 0.0084 & 0.0105 & 0.2218 &0.1640 
    & 0.2717 &0.1598 &1.1498 &0.3551
    & 0.0087 &0.0033 &0.1160 &0.0342
    & 0.0099 &0.0045 &0.3953 &0.2794  \\
\bottomrule
\end{tabular}
}\\
\label{tab:mean_vars}
\end{table*}

\begin{figure*}[ht]
  \centering
  \includegraphics[width=0.91\linewidth]{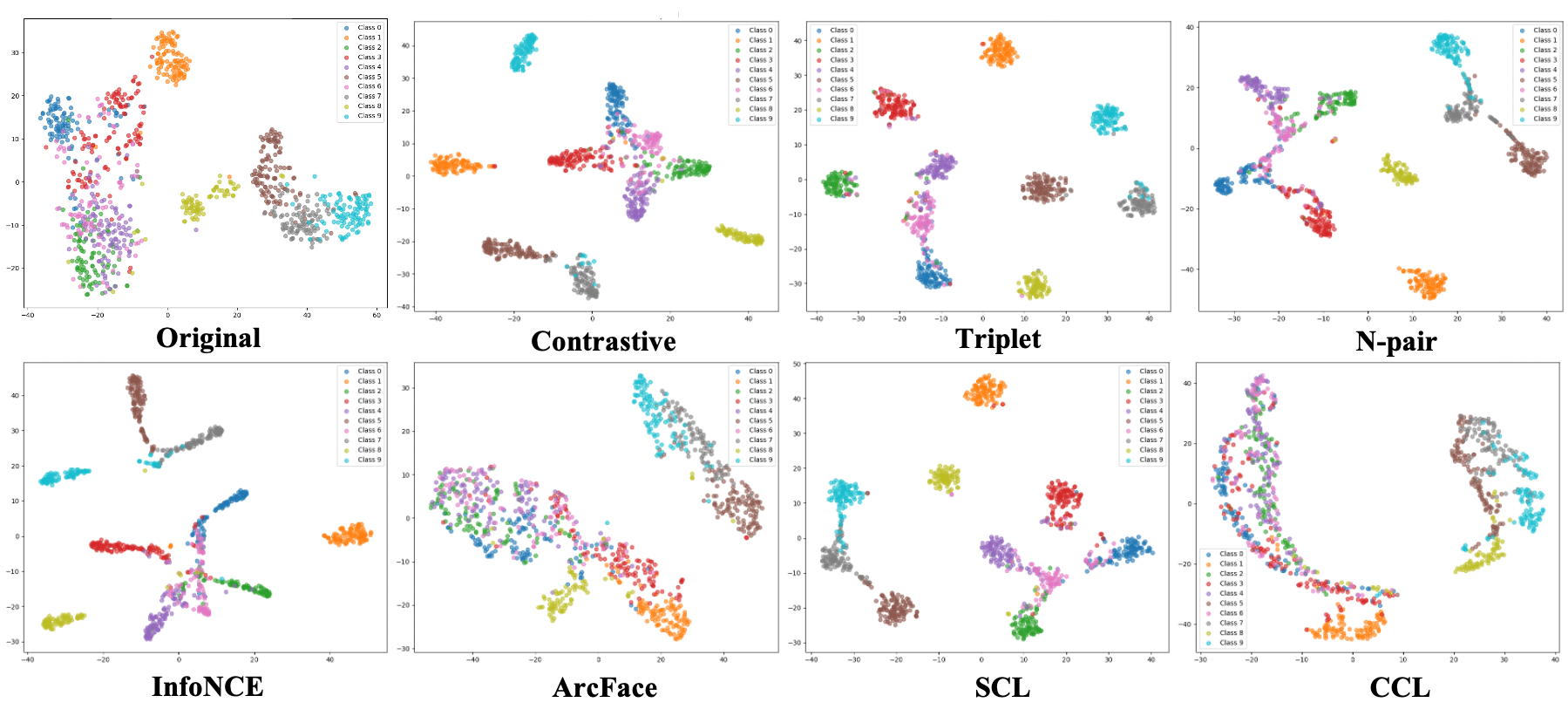}
  \caption{t-SNE projection of final embeddings on FashionMNIST (random subset of 1,000 samples), each color denotes a class.}
  \label{fig:t_sne}
\end{figure*}

\section{Results and Analysis}
\subsection{Variance: Quantifying Geometric Granularity}
We employ VARIANCE diagnostics to quantify the geometric granularity of the resulting embedding spaces. Table~\ref{tab:mean_vars} reports per-loss intra- and inter-class statistics (means and variances) across five datasets. Several clear patterns emerge. Triplet-based training consistently preserves larger within-class variance while producing substantial inter-class separation, supporting fine-grained distinctions (e.g., CIFAR-10: triplet intra=0.1435 vs contrastive intra=0.0656, triplet inter=1.3653). Contrastive objectives tend to compact intra-class embeddings (small intra- means) and yield more modest class-center separation. N-pair and InfoNCE often achieve large inter-class means (strong centroid separation) but differ in intra behavior: N-pair typically yields compact intra clusters with high inter/intra ratios, while InfoNCE is more dataset-dependent. Methods that explicitly enforce class-centers (CCL, ArcFace) produce extremely small intra- means in our measurements; however, ArcFace shows zero intra- means on several datasets (Table~\ref{tab:mean_vars}), indicating a metric/scale mismatch (angular vs. cosine/Euclidean) and should be interpreted with care. Finally, inter-class variances vary substantially across losses (e.g., N-pair inter $\sigma^2$ is large on Tiny-ImageNet), indicating heterogeneity in class spacing that can affect retrieval: large mean separation alone does not guarantee uniformly good nearest-neighbor behavior.
Fig.~\ref{fig:t_sne} shows T-SNE projections of the learned embeddings. Contrastive training produces compact, tightly clustered class groups; Triplet and SCL yield more dispersed within-class clusters while maintaining clear class separation, indicating higher preserved intra-class variance. These qualitative patterns are consistent with the VARIANCE diagnostics reported in Table~\ref{tab:mean_vars} and Fig.~\ref{fig:t_sne}.

\begin{table*}[ht]
\small
\centering
\setlength{\tabcolsep}{2pt} 
\caption{Recall@k ($k=1,5,10$, abbreviated as 'r@k') for image retrieval across five benchmark datasets.}
\begin{tabular}{lcccccccccccccccccc}
\toprule
\multirow{2}{*}{Loss} & \multicolumn{3}{c}{CIFAR-10}  & \multicolumn{3}{c}{CARS196} &\multicolumn{3}{c}{CUB-200} &\multicolumn{3}{c}{Tiny-ImageNet } &\multicolumn{3}{c}{FashionMNIST} \\ 
\cmidrule(lr){2-4} \cmidrule(lr){5-7} \cmidrule(lr){8-10} \cmidrule(lr){11-13} \cmidrule(lr){14-16} 
& r@1 & r@5 & r@10 
& r@1 & r@5 & r@10 
& r@1 & r@5 & r@10 
& r@1 & r@5 & r@10 
& r@1 & r@5 & r@10\\
\midrule
Contrastive 
    & 0.9248  &0.9742 &0.9831
    & 0.3164 &0.6027 & 0.7162 
    & 0.5138 &0.7594 &0.8410 
    &0.7030  &0.8120  & 0.8457
    & 0.8848 &0.9653 &0.9800 \\
Triplet     
    & 0.9219 &0.9753 &0.9827
    & 0.3174 &0.5996 &0.7228
    & 0.4641 &0.7308 &0.8234
    & 0.7303 & 0.8486 & 0.8799
    & 0.8903 & 0.9635 &0.9749 \\  
N-pair        
    & 0.9134 &0.9724 &0.9813
    & 0.1087 &0.2645 &0.3953
    & 0.2192 &0.4791 &0.6120
    &0.4367 &0.6578 &0.7506 
    & 0.8487 &0.9543 &0.9713 \\
InfoNCE       
    & 0.9228 &0.9735 &0.9820
    & 0.2364 &0.5084 &0.6305
    & 0.4610 &0.7040 &0.7915
    &0.7416 &0.8364 &0.8630
    & 0.8754 &0.9615 &0.9752 \\
ArcFace       
    &  0.7781 &0.9313 &0.9617
    & 0.1301 &0.3262 &0.4511
    & 0.2972 & 0.5609 &0.6814
    & 0.4309 & 0.6229 & 0.7063 
    & 0.6948 & 0.9121 & 0.9589 \\
SCL       
    & 0.9272 &0.9729 &0.9820
    & 0.4285 &0.7002 &0.7913
    & 0.4408  &0.7076 &0.8029
    & 0.6486 &0.8173 &0.8521
    & 0.8813 &0.9654 &0.9783 \\
CCL       
    & 0.8340 &0.9487 &0.9693
    & 0.1204 &0.3151 &0.4328
    & 0.3148 &0.5754 &0.6945
    & 0.6846 & 0.8294 & 0.8728  
    & 0.7132 &0.9187 &0.9625 \\
\bottomrule
\end{tabular}
\vspace{-15pt}
\label{tab:retrieval_result}
\end{table*}

\subsection{Greediness: Characterizing Optimization Efficiency}
\label{sec:greediness_in_optimization}
\begin{figure}[h]
    \centering
    \includegraphics[width=0.95\linewidth]{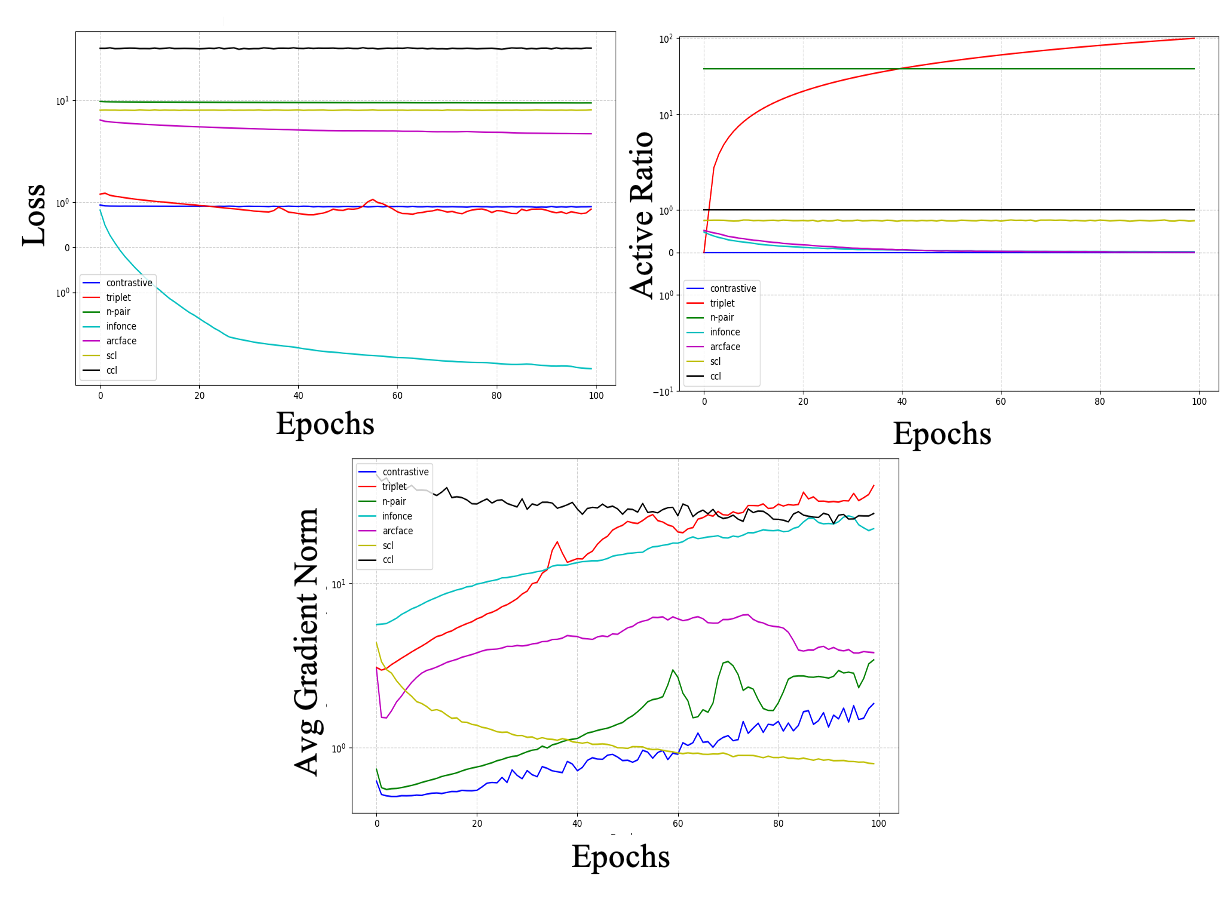}
    \caption{Greediness metrics (symlog scale~\cite{symlog}) over training epochs on CARS196 dataset.}
    \label{fig:greediness_metrics}
\end{figure}

To understand the "cost" of achieving these geometric structures, we use GREEDINESS metrics to characterize optimization efficiency. Contrastive-style objectives (InfoNCE-like) behave as \textit{greedy \& diffuse learners}: they reduce training loss rapidly ($\approx$ 50\% reduction by epoch 30 in our runs), engage a large fraction of samples early (active ratio $\approx$ 65\%), and produce many small gradient updates (average norm $\approx$ 0.12). These dynamics drive quick compaction of intra-class distances and early gains in retrieval on coarse datasets, but they also tend to plateau on hard examples and can obscure fine-grained distinctions. By contrast, batch-hard triplet (SCL exhibits similar) is \textit{focused \& persistent}: it reaches the 60\% loss-reduction point much later ($\approx$ epoch 40), uses a substantially smaller active set ($\approx$ 38\%) and larger gradients (average norm  $\approx$ 0.27). Those fewer but stronger updates concentrate learning on hard samples, preserve higher within-class variance, and correlate with improved fine-grained retrieval performance, illustrated in Figure~\ref{fig:greediness_metrics} and Table~\ref{tab:mean_vars}.
These diagnostics provide actionable practical guidance. High mean inter-class separation, as seen in N-pair and certain contrastive variants, can mask uneven centroid spacing; thus, mean separation alone does not guarantee a uniform nearest-neighbor manifold. Furthermore, extreme intra-class compaction in angular or center-based methods (e.g., ArcFace, CCL) often reflects metric or normalization mismatches rather than superior clustering and requires cautious interpretation. Ultimately, these findings reveal a fundamental efficiency-granularity trade-off. We recommend reporting GREEDINESS alongside VARIANCE metrics to navigate this balance: prioritize Triplet/SCL when intra-class diversity and hard-sample discrimination are critical, and select Contrastive/InfoNCE when optimization efficiency and rapid embedding refinement are the primary objectives.    

\subsection{Retrieval Performance Across Datasets}
Table~\ref{tab:retrieval_result} shows Recall@k across five datasets. No single loss dominates everywhere: Contrastive leads on CIFAR-10 and CUB-200, Triplet excels on Tiny-ImageNet and FashionMNIST, and SCL achieves the best R@1 on CARS196. InfoNCE is consistently competitive, while N-pair struggles on fine-grained datasets. ArcFace and CCL underperform overall, consistent with their near-collapsed intra-class statistics (Table~\ref{tab:mean_vars}). These trends align with our variance–greediness analysis. Triplet and SCL preserve larger intra-class variance, which benefits top-1 retrieval in fine-grained or noisy datasets. Contrastive and InfoNCE compact clusters more aggressively, which helps on simpler datasets like CIFAR-10 but risks obscuring subtle distinctions. Thus, the choice of loss should match task demands: Triplet/SCL for fine-grained retrieval and Contrastive/InfoNCE for faster, compact embedding refinement.
\vspace{-10pt}
\section{Conclusion}
Our work moved beyond standard performance leaderboards to provide a fundamental characterization of how metric-learning objectives govern representation learning. Using our VARIANCE and GREEDINESS framework, we identified a critical trade-off between optimization efficiency and geometric preservation: while Contrastive and InfoNCE losses offer rapid cluster compaction, they often sacrifice the intra-class diversity that Triplet and SCL losses inherently maintain. Furthermore, our findings regarding N-pair loss highlight the danger of relying on aggregate distance metrics alone, as they can mask underlying structural irregularities in the embedding space. These results suggest that the "optimal" loss is not a fixed target but a choice dictated by whether a task requires fine-grained discrimination or coarse-grained efficiency. Ultimately, our work establishes a diagnostic roadmap for researchers to navigate these dynamics, serving as a vital complement to standard benchmarks. 

\bibliography{refs}
\bibliographystyle{plain}
\end{document}